\documentclass[runningheads,a4paper]{llncs}

\usepackage{amssymb}
\usepackage{graphicx}
\usepackage{times}

\usepackage{comment} %lan add
\usepackage{graphicx} %lan add
\usepackage{epstopdf} %lan add
\usepackage{amsmath} %zhu add
\usepackage{mathtools} % zhu add
\usepackage{mathrsfs} %zhu add
\usepackage{scrextend} %zhu add
\usepackage{url} %lan add
\usepackage{epstopdf}%lan add
\usepackage{bm} %lan add
\usepackage{amssymb} %lan add
\usepackage{color}
\usepackage{flushend} %lan add
\makeatletter
\def\hlinew#1{%
	\noalign{\ifnum0=`}\fi\hrule \@height #1 \futurelet
	\reserved@a\@xhline}
\makeatother

\newcommand{\tabincell}[2]{\begin{tabular}{@{}#1@{}}#2\end{tabular}} %zhu add
\usepackage{url}
\newcommand{\keywords}[1]{\par\addvspace\baselineskip
\noindent\keywordname\enspace\ignorespaces#1}

\begin{document}

\mainmatter  % start of an individual contribution

\title{Adversarial Deep Structural Networks for Mammographic Mass Segmentation}

\titlerunning{Adversarial Deep Structural Networks for Mammographic Mass Segmentation.}

\author{Wentao Zhu \and Xiang Xiang \and Trac~D.~Tran \and Xiaohui Xie} 
% index{Zhu, Wentao}
% index{Lou, Qi}
% index{Vang, Yeeleng}
% index{Xie, Xiaohui}
%\thanks{Please note that the LNCS Editorial assumes that all authors have used
%the western naming convention, with given names preceding surnames. This determines
%the structure of the names in the running heads and the author index.}%
%\and Ursula Barth\and Ingrid Haas\and Frank Holzwarth\and\\
%Anna Kramer\and Leonie Kunz\and Christine Rei\ss\and\\
%Nicole Sator\and Erika Siebert-Cole\and Peter Stra\ss er}
%
\authorrunning{W. Zhu et al.}
%\institute{Anonymous Authors}
%\institute{Paper ID: 415}
\institute{University of California, Irvine and Johns Hopkins University \\ \{wentaoz1, xhx\}@ics.uci.edu, \{xxiang, trac\}@jhu.edu}
%\institute{University of California, Irvine}
%\institute{\{wentaoz1, xhx\}@ics.uci.edu, \{qlou, ysvang\}@uci.edu}
%\institute{Springer-Verlag, Computer Science Editorial,\\
%Tiergartenstr. 17, 69121 Heidelberg, Germany\\
%\url{http://www.springer.com/lncs}}
\toctitle{Lecture Notes in Computer Science}
\tocauthor{Authors' Instructions}
\maketitle
\begin{abstract}
  Mass segmentation is an important task in mammogram analysis, providing effective morphological features and regions of interest (ROI) for mass detection and classification. Inspired by the success of using deep convolutional features for natural image analysis and conditional random fields (CRF) for structural learning, we propose an end-to-end network for mammographic mass segmentation. The network employs a fully convolutional network (FCN) to model a potential function, followed by a CRF to perform structured learning. Because the mass distribution varies greatly with pixel position, the FCN is combined with a position priori. Due to the small size of mammogram datasets, we use adversarial training to control over-fitting. Four models with different convolutional kernels are further fused to improve the segmentation results. Experimental results on two public datasets, INBreast and DDSM-BCRP, demonstrate that our end-to-end network achieves the state-of-the-art results. \footnote[1]{Code: https://github.com/wentaozhu/adversarial-deep-structural-networks.git.}
\end{abstract}
\section{Introduction}\label{sec:intro}
According to the American Cancer Society, breast cancer is the most frequently diagnosed solid cancer and the second leading cause of cancer death among U.S. women~\cite{acs}. Mammogram screening has been demonstrated to be an effective way for early detection and diagnosis, which can significantly decrease breast cancer mortality \cite{oeffinger2015breast}. Mass segmentation provides morphological features, which play crucial roles for diagnosis.

Traditional studies on mass segmentation rely heavily on elaborate human designed features. Model-based methods build classifiers and learn the features from the mass \cite{beller2005example,cardoso2015closed}. There are few works using deep networks to process the mammogram~\cite{kallenberg2016unsupervised,greenspan2016guest,zhu2016deep}. Dhungel et al. employed multiple deep belief networks (DBNs), GMM classifier and a priori as potential functions, and structural SVM to perform segmentation \cite{dhungel2015deep}. They also used CRF with tree re-weighted belief propagation to boost the segmentation results \cite{dhungel2015tree}. A recent work used the output from a CNN as a complimentary potential function, yielding the state-of-the-art performance~\cite{dhungel2015deepmiccai}. However, the two-stage training used in these methods produces potential functions that can easily over-fit training data.% and cannot handle spatial information that might be important for segmentation in CNN. %Contour-based methods typically need complex image processing or mathematical models to detect boundaries using methods such as level sets \cite{cardoso2015closed}.

\begin{figure}[t]
	\begin{center}
		%\begin{tabular}{cc}
		\begin{minipage}{0.4\linewidth}
			\centerline{\includegraphics[width=6cm]{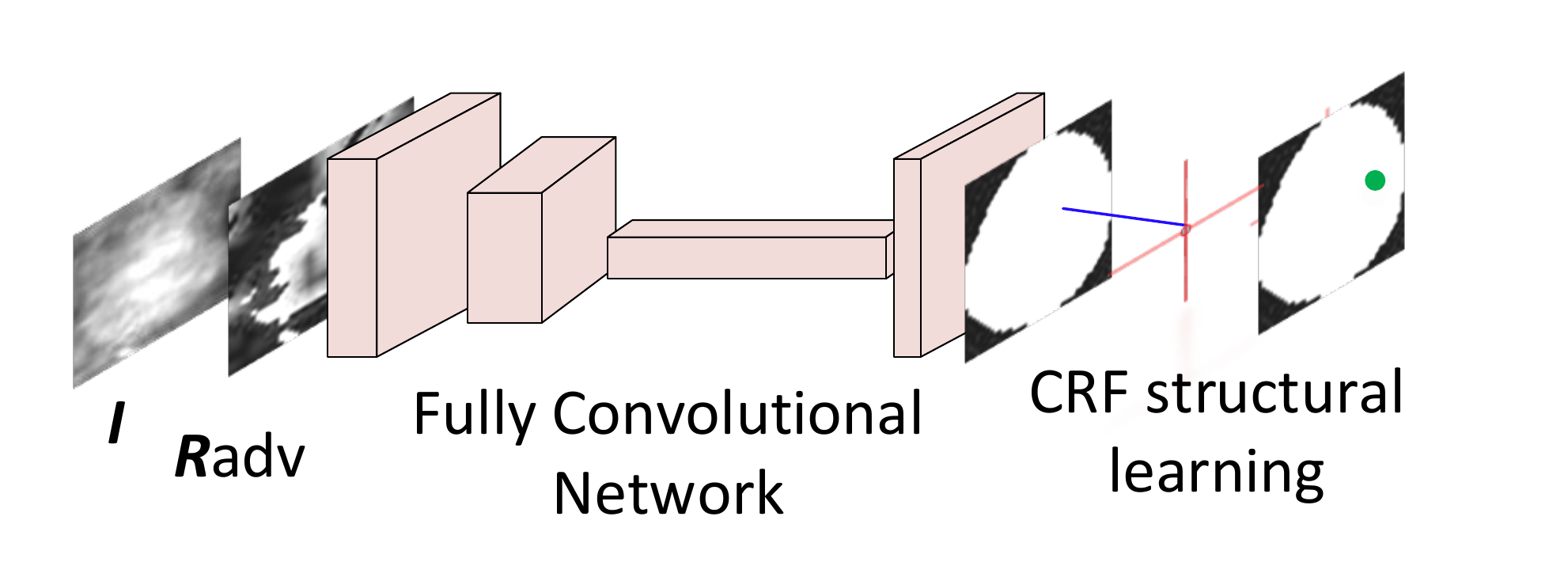}}
			\center{(a)}
		\end{minipage}
		\hspace{1cm}
		\begin{minipage}{0.4\linewidth}
			\centerline{\includegraphics[width=7.5cm]{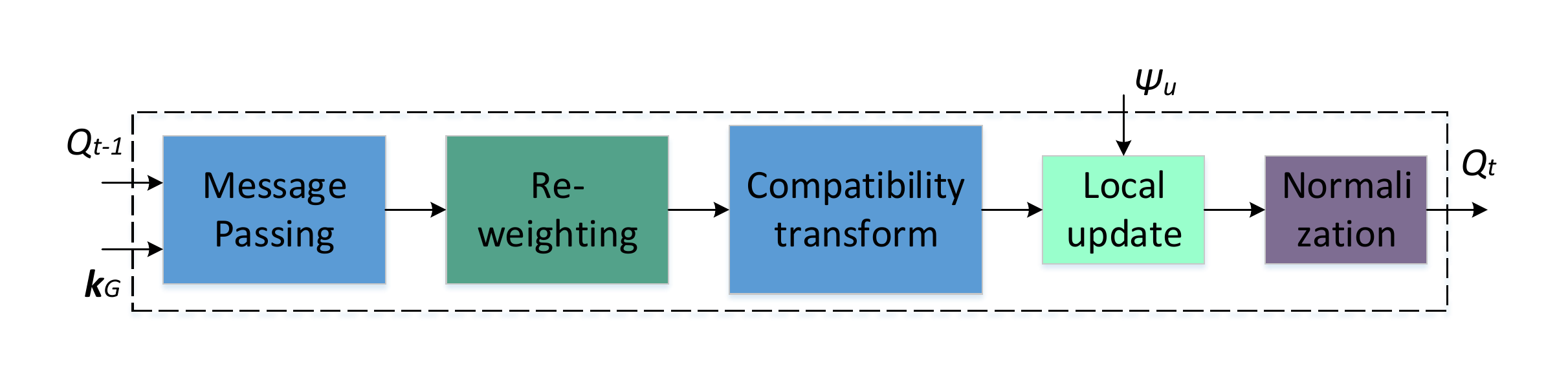}}
			\center{(b)}
		\end{minipage}
		\caption{The proposed adversarial deep FCN-CRF network with four convolutional layers followed by CRF structured learning (a). The recurrent unit in the CRF as RNN, which consists of message passing, re-weighting, compatibility transform, local update and normalization (b).}
		\label{framework}
	\end{center}
\end{figure}

Inspired by the power of deep networks~\cite{zhu2016co,zhu2015hierarchical}, we propose an end-to-end trained adversarial deep structural network to perform mass segmentation (Fig. \ref{framework}(a)). The proposed network is designed to robustly learn from a small dataset with poor contrast mammographic images. Specifically, an end-to-end trained fully convolution network (FCN) with CRF is applied. Adversarial training is introduced into the network to learn robustly from scarce mammographic images. To further explore statistical property of mass regions, the spatial priori with categorical distributions considering the positions are added into FCN. We validate our adversarial deep structural network on two public mammographic mass segmentation datasets. The proposed network is demonstrated to consistently outperform other algorithms for mass segmentation. %learn from the poor contrastive and scarce mammographic images With these components,

Our main contributions in this paper are: (1) It is the first time to apply adversarial training to medical imaging. Integrating CNN+CRF and adversarial training into a unified end-to-end training framework has not been attempted before. Both components are essential and necessary for achieving the state-of-the-art performance. (2) We employ an end-to-end trained network to do the mass segmentation while previous works needed a lot of hand-designed features or multi-stage training, such as calculating potential functions independently. (3) Our model achieves state-of-the-art results on two commonly used mammographic mass segmentation datasets.%, the Inbreast dataset and the DDSM-BCRP dataset.

\section{Fully Convolution Network-CRF Network}
Fully convolutional network (FCN) is a successful model for image segmentation, which preserves the spatial structure of predictions~\cite{long2015fully}. FCN consists of convolution, de-convolution \cite{zeiler2010deconvolutional}, or max-pooling in each layer. For training, the FCN optimizes maximum likelihood loss function $\mathcal{L}_{FCN} = -\frac{1}{N\times N_i} \sum_{n=1}^{N}\sum_{i=1}^{N_i} \log p_{fcn}{(y_{n,i} | \textbf{I}_n, \bm \theta)}$, where $y_{n,i}$ is the label of $i$th pixel in the $n$th image $\textbf{I}_n$, $N$ is the number of training mammograms, $N_i$ is the number of pixels in the image, and $\bm \theta$ is the parameter of FCN. Here the size of images is fixed to $40 \times 40$ and $N_i$ is 1,600. %\begin{equation}
%\label{equ:fcnloss}
%\mathcal{L}_{FCN} = -\frac{1}{N} \sum_{n=1}^{N}\sum_{i=1}^{N_i} \log p(\textbf{I}_n | y_{n,i}, \bm \theta),
%\end{equation}

CRF is a commonly used method for structural learning, well suited for image segmentation. It models pixel labels as random variables in a Markov random field conditioned on an observed input image. To make the annotation consistent, we use $\textbf{y} = (y_1, y_2, \dots, y_i, \dots, y_{1600})^T$ to denote the random variables of pixel labels in an image, where $y_i\in\{0,1\}$. The zero denotes pixel belonging to background, and one denotes it belonging to mass region. The Gibbs energy of fully connected pairwise CRF \cite{krahenbuhl2011efficient} is $E(\textbf{y}) = \sum_{i} \psi_u (y_i) + \sum_{i<j} \psi_p (y_i, y_j)$,
where unary potential function $\psi_u (y_i)$ is the loss of FCN in our case, pairwise potential function $\psi_p (y_i, y_j)$ defines the cost of labeling pair $(y_i, y_j)$. The pairwise potential function $\psi_p (y_i, y_j)$ can be defined as 
\begin{equation}
\label{equ:pairwisepotential}
\psi_p (y_i, y_j) = \mu (y_i, y_j) \sum_{m} w^{(m)} k_G^{(m)}(\textbf{f}_i, \textbf{f}_j),
\end{equation}
where label compatibility function $\mu$ is given by the Potts model in our case, $k_G^{(m)}$ is the Gaussian kernel applied to feature vectors \cite{krahenbuhl2011efficient}, $w^{(m)}$ is the learned weight. Pixel values $I_i$ and positions $p_i$ can be used as the feature vector $\textbf{f}_i$. %\begin{equation}
%\label{equ:crfenergy}
%E(\textbf{y}) = \sum_{i} \psi_u (y_i) + \sum_{i<j} %\psi_p (y_i, y_j),
%\end{equation}

Efficient inference algorithm can be obtained by mean field approximation $Q(\textbf{y}) = \prod_i Q_i(y_i) $ \cite{krahenbuhl2011efficient}. The update rule is
\begin{equation}
\label{equ:messagepassing}
\begin{aligned}
&\tilde{Q}_i^{(m)}(l)\leftarrow \sum_{i\neq j}k_G^{(m)}(\textbf{f}_i,\textbf{f}_j)Q_j(l) \text{ for all } m,\\
&\check{Q}_i(l)\leftarrow \sum_m w^{(m)}\tilde{Q}_i^{(m)}(l), \quad \hat{Q}_i(l)\leftarrow \sum_{l^\prime\in\mathcal{L}}\mu(l,l^\prime)\check{Q}_i(l),\\
&\breve{Q}_i(l)\leftarrow \exp(- \psi_u (y_i = l))-\hat{Q}_i(l), \quad Q_i\leftarrow \frac{1}{Z_i}\exp\left(\breve{Q}_i(l)\right),
\end{aligned}
\end{equation}
where the first line is the message passing from label of pixel $i$ to label of pixel $j$, the second line is re-weighting with the learned weights $w^{(m)}$, the third line is compatibility transform, the fourth line is adding unary potentials, and the last step is normalization operator. Here $\mathcal{L} = \{0,1\}$ denotes background or mass. The initialization of inference employs unary potential function as $Q_i(y_i) = \frac{1}{Z_i} \exp(- \psi_u (y_i))$. The above mean field approximation can be interpreted as a recurrent neural network (RNN) in Fig.~\ref{framework}(b)\cite{zheng2015conditional}. %The recurrent unit takes last state's approximation $Q_{t-1}$, unary potential function $\psi_u$, and the Gaussian kernel $\textbf{k}_G$ as the input. The output is updated approximation $Q_{t}$. After interpreting the CRF as a RNN, the framework using FCN as potential function followed by CRF can be trained with an end-to-end scheme \cite{zheng2015conditional}.
\section{Adversarial FCN-CRF Network}
\begin{figure}[t]
	\begin{center}
		%\begin{tabular}{cc}
		\begin{minipage}{0.3\linewidth}
			\centerline{\includegraphics[width=1cm]{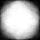}	\includegraphics[width=1cm]{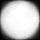}}
			\center{(a)}
		\end{minipage}
		\hspace{1cm}
		\begin{minipage}{0.6\linewidth}
			\centerline{
			\includegraphics[width=1cm]{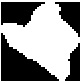}
			\includegraphics[width=1cm]{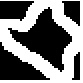}
			\includegraphics[width=1cm]{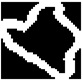}}
			\center{(b)}
		\end{minipage}
		\caption{The empirical estimation of a priori on INBreast (left) and DDSM-BCRP (right) training datasets (a). Trimap visualizations on the DDSM-BCRP dataset, segmentation groundtruth (first column), trimap of width $2$ (second column), trimaps of width $3$ (third column) (b).}
		\label{prior}
	\end{center}
\end{figure}

Shape and the appearance priori play an important role in mammogram mass segmentation \cite{jiang2016mammographic,dhungel2015deepmiccai}. The distribution of labels varies greatly with a position in the mammographic mass segmentation. From observation, most of the mass is located in the center of ROI, and the boundary of ROI is more likely to be background (Fig. \ref{prior}(a)).% which is validated by the result that priori alone gives much better segmentation than multiple DBNs and GMM classifier \cite{dhungel2015deepmiccai}.

The conventional FCN provides predictions for pixels independently. It only considers global class distribution difference corresponding to the number of filters (channels) in the last layer. Here we take the categorical priori of different positions into consideration and add it into the FCN as $p(y_i | \textbf{I}, \bm \theta) \propto p(y_i)p_{fcn}{(y_i | \textbf{I}, \bm \theta)}$, where $p(y_i)$ is the categorical priori distribution varied with the pixel position $i$, and $p_{fcn}{(y_i | \textbf{I}, \bm \theta)}$ is the output of conventional FCN. In the implementation, we assigned the bias of last layer as the average image to train network. The $-\log p(y_i | \textbf{I}, \bm{\theta})$ is used as the unary potential function for $\psi_u (y_i)$ in the CRF as RNN. For multiple FCNs as potential functions, the potential function is defined as $\psi_u (y_i) = \sum_{u \prime} w_{(u \prime)} \psi_{u \prime}(y_i)$, where $w_{(u \prime)}$ is the learned weight for unary potential function, $\psi_{u \prime}(y_i)$ is the  potential function provided by one FCN. %Experimental result shows that adding the position priori obtains 0.25\% Dice index improvement than the result of FCN without the position priori on the INBreast dataset.% \cite{moreira2012inbreast}. %\begin{equation}
%\label{equ:prior}
%p(y_i | \textbf{I}, \bm \theta) \propto %p(y_i)p(\textbf{I} | y_i, \bm \theta),
%\end{equation}
%\begin{equation}
%\label{equ:multipotential}
%\psi_u (y_i) = \sum_{u \prime} w_{(u \prime)} \psi_{u \prime}(y_i),
%\end{equation}

Adversarial training provides strong regularization for deep networks \cite{goodfellow2014explaining}. The idea of adversarial training is that if the model is robust enough, it should be invariant to small perturbations of training examples that yield the largest increase in the loss (adversarial examples \cite{szegedy2013intriguing}). The perturbation $\bm{R}$ for adversarial example can be obtained as $\min_{\textbf{R}, \| \textbf{R} \| \leq \epsilon} \log p(\bm{y} | \textbf{I} + \textbf{R}, \bm \theta)$. In general, the calculation of exact $\bm R$ is intractable because the exact minimization is not solvable w.r.t. $\bm R$, especially for complicated models such as deep networks. The linear approximation and $L_2$ norm box constraint can be used for the calculation of perturbation \cite{goodfellow2014explaining} as $\bm{R}_{adv} = - \frac{\epsilon \bm{g}}{\|\bm{g}\|_2}$,
where $\bm{g} = \nabla_{\bm{I}} \log p(\bm{y} | \bm{I}, \bm{\theta})$. For adversarial fully convolutional network, the network predicts label of each pixel independently as $p(\bm{y}| \bm{I}, \bm{\theta}) = \prod_{i} p(y_i | \bm{I}, \bm{\theta})$. For adversarial CRF as RNN, the prediction of network relies on mean field approximation inference as $p(\bm{y}| \bm{I}, \bm{\theta}) = \prod_{i} Q(y_i | \bm{I}, \bm{\theta})$.%\begin{equation}
%\label{equ:adversarialpert}
%\bm{R}_{adv} = - \frac{\epsilon \bm{g}}{\|\bm{g}\|_2},
%\end{equation}
%\begin{equation}
%\label{equ:adversarialtrain}
%\min_{\textbf{R}, \| \textbf{R} \| \leq \epsilon} \log p(\bm{y} | \textbf{I} + \textbf{R}, \bm \theta).
%\end{equation}

The adversarial training forces the model to fit examples with the worst perturbation as well. The adversarial loss is defined as 
\begin{equation}
\label{equ:adversarialloss}
\mathcal{L}_{adv}(\bm{\theta}) = - \frac{1}{N} \sum_{n=1}^{N} \log p(\bm{y}_n | \bm{I}_n + \bm{R}_{adv,n}, \bm{\theta}).
\end{equation}
In training, the total loss is defined as the sum of adversarial loss and the empirical loss based on training samples as 
\begin{equation}
\label{equ:totalloss}
\mathcal{L}(\bm{\theta}) = \mathcal{L}_{adv}(\bm{\theta})- \frac{1}{N} \sum_{n=1}^{N} \log p(\bm{y}_n | \bm{I}_n, \bm{\theta}) + \frac{\lambda}{2} \| \bm{\theta} \|^2,
\end{equation}
where $\lambda$ is the regularization factor used to avoid over-fitting, $p(\bm{y}_n | \bm{I}_n, \bm{\theta})$ is either prediction in the enhanced FCN or posteriori approximated by mean field inference in the CRF as RNN for the $n$th image $\bm{I}_n$. The $L_2$ regularization term is used only for the parameters in CRF.
\section{Experiments}\label{sec:exp}
We validate the proposed model on two publicly and most frequently used mammographic mass segmentation datasets: INBreast dataset \cite{moreira2012inbreast} and DDSM-BCRP dataset \cite{heath1998current}. We use the same ROI extraction and re-size principle as \cite{dhungel2015deep,dhungel2015deepmiccai,dhungel2015tree}. Due to the low contrast of mammographic images, image enhancement technique is used on the extracted ROI images as the first 9 steps of enhancement \cite{ball2007digital}, followed by pixel position dependent normalization. The preprocessing makes training converge quickly. We further augment each training set by flipping horizontally, flipping vertically, flipping horizontally and vertically, which makes the training set 4 times larger than the original training set. 

%$0.9013$ 0.9004
For consistent comparison, the Dice index metric is used for the segmentation results and is defined $\frac{2 \times TP}{2 \times TP + FP + FN}$. For a fair comparison, we also validate the Deep Structure Learning + CNN \cite{dhungel2015deepmiccai} on our processed data, and obtain similar result (Dice index $0.9010$) on the INBreast dataset. To investigate the impact of each component in our model, we conduct extensive experiments under different configurations. FCN is the network integrating a position priori into FCN (structure denoted as FCN 1 in Tab. 1). We use the enhanced FCN rather than the conventional FCN in all experiments. Adversarial FCN is FCN  with adversarial training. Jointly Trained FCN-CRF is the FCN followed by CRF as RNN with an end-to-end training scheme. Jointly Trained Adversarial FCN-CRF is the Jointly Trained FCN-CRF with end-to-end adversarial training. Multi-FCN, Adversarial Multi-FCN, Jointly Trained Multi-FCN-CRF, Jointly Trained Adversarial Multi-FCN-CRF are those networks with 4 FCNs. The configuration of FCN and other used three subnetworks in the Multi-FCN are in Table~\ref{tab:network}. The last layers of the four networks are all two $40\times40$ deconvolutional filters with softmax activation function. We use hyperbolic tangent activation function in middle layers. The parameters of FCNs are set such that the number of each layer's parameters is almost the same as that of CNN used in the work \cite{dhungel2015deepmiccai}. For optimization, we use Adam algorithm \cite{kingma2014adam} with learning rate 0.003. The $\lambda$ used for weights of CRF as RNN is $0.5$ in the two datasets. The $\epsilon$ used in adversarial training are $0.1$ and $0.5$ for INBreast and DDSM-BCRP datasets respectively, because the boundaries of masses on the DDSM-BCRP dataset are smoother than those on the INbreast dataset. For mean field approximation or the CRF as RNN, we use 5 iterations/time steps in the training and 10 iterations/time steps in the test phase. %Due to the scarcity of data, we simplify the Gaussian kernel $\bm{k}_G(\textbf{f}_i, \textbf{f}_j)$ by only considering the neighboring pixels
%\begin{equation}
%\label{equ:gaussiankernel}
%\bm{k}_G(\textbf{f}_i, \textbf{f}_j) = [\exp(-|I_i - I_j|^2/2), \exp(-|p_i - p_j|^2)/2]^T,
%\end{equation}
%which is a special form of truncated Gaussian kernel.
\begin{table}[t]
	\fontsize{9pt}{10pt}\selectfont\centering
	\caption{Configurations of sub-networks in the Multi-FCN.}\label{tab:network}
	\begin{tabular}{c|c|c|c}
		\hlinew{0.9pt}
		Net.&First layer&Second layer&Third layer\\
		\hline
		FCN 1&$6\times5\times5$ conv., $2\times2$ max pool.&$12\times5\times5$ conv., $2\times2$ max pool.&$588\times7\times7$ conv.\\
		\hline
		FCN 2&$9\times4\times4$ conv., $2\times2$ max pool.&$12\times4\times4$ conv., $2\times2$ max pool.&$588\times7\times7$ conv.\\
		\hline
		FCN 3&$16\times3\times3$ conv., $2\times2$ max pool. & $13\times3\times3$ conv., $2\times2$ max pool.&$415\times8\times8$ conv.\\
		\hline
		FCN 4&$37\times2\times2$ conv., $2\times2$ max pool.&$12\times2\times2$ conv., $2\times2$ max pool.&$355\times9\times9$ conv.\\
		\hlinew{0.9pt}
	\end{tabular}
\end{table}

\begin{table}[t]
	\fontsize{9pt}{10pt}\selectfont\centering
	\caption{Comparisons on INBreast and DDSM-BCRP datasets.}\label{tab:inbreast}
	\begin{tabular}{c|c|c}
		\hlinew{0.9pt}
		Methodology&INBreast Dice(\%)&DDSM-BCRP Dice(\%)\\		
		\hlinew{0.7pt}
		\tabincell{c}{Cardoso et al. \cite{cardoso2015closed}}&88&N/A \\
		\hline
		\tabincell{c}{Beller et al. \cite{beller2005example}}&N/A&70\\
		\hline
		\tabincell{c}{Deep Structure Learning \cite{dhungel2015deep}}&88&87\\
		\hline
		\tabincell{c}{TRW Deep Structure Learning \cite{dhungel2015tree}}&89&89\\
		\hline
		\tabincell{c}{Deep Structure Learning + CNN \cite{dhungel2015deepmiccai}}&90&90\\
		\hlinew{0.9pt}
		FCN & 89.48 & 90.21 \\
		\hline
		\tabincell{c}{ FCN with Adversarial Training} & 89.71 & 90.78\\
		\hline
		\tabincell{c}{Jointly Trained FCN-CRF} & 89.78 & 90.97\\
		\hline
		\tabincell{c}{ FCN-CRF with Adversarial Training}& 90.07 & 91.03\\
		\hline
		\tabincell{c}{Multi-FCN} & 90.47 & 91.17\\
		\hline
		\tabincell{c}{Multi-FCN with Adversarial Training} & 90.71 & 91.20\\
		\hline
		\tabincell{c}{Jointly Trained Multi-FCN-CRF} & 90.76 & 91.26\\
		\hline
		\tabincell{c}{Multi-FCN-CRF with Adversarial Training} & \textbf{90.97} & \textbf{91.30}\\
		\hlinew{0.9pt}
		\end{tabular}
\end{table}
The INBreast dataset is a recently released mammographic mass analysis dataset, which provides more accurate contours of lesion region and the mammograms are of high quality. For mass segmentation, the dataset contains 116 mass regions. We use the first 58 masses for training and the rest for test, which is of the same protocol used in these works \cite{dhungel2015deep,dhungel2015deepmiccai,dhungel2015tree}. The DDSM-BCRP dataset contains 39 cases (156 images) for training and 40 cases (160 images) for testing~\cite{heath1998current}. After ROI extraction, there are 84 ROIs for training, and 87 ROIs for test. We compare our schemes with other recently published mammographic mass segmentation methods \cite{cardoso2015closed,dhungel2015deep,dhungel2015tree,dhungel2015deepmiccai} in Table \ref{tab:inbreast}.

%about 0.5\% improvement compared to a CNN followed by a fully connected network (88.98\%
Table \ref{tab:inbreast} shows that the successfully used CNN features in natural image provide superior performance on medical image analysis, outperforming hand-crafted feature based methods \cite{cardoso2015closed,beller2005example}. Our enhanced FCN achieves 0.25\% Dice index improvement than the traditional FCN on the INBreast dataset. The adversarial training yields 0.4\% improvement on average. Incorporating the spatially structural constraint further produces 0.3\% improvement. Using model average or multiple potential functions contributes the most to segmentation results which is consistent with work showing that the best model requires five different unary potential functions \cite{dhungel2015deepmiccai}. Combining all the components together achieves the best performance with relative 9.7\%, 13\% improvement on INBreast, DDSM-BCRP datasets respectively. In our experiment, the FCN overfits heavily on the training set and can even achieve above 98.60\% Dice index. It might explains why the two-stage training cannot boost the performance too much. The adversarial training works effectively as a regularization to reduce the overfitting. We believe that the overfitting is mainly caused by the small training set size and we strongly support the creation of a large mammographic analysis dataset to accelerate mammogram analysis research.

We calculate the p-value of McNemar’s Chi-Square Test to compare our model with the method~\cite{dhungel2015deepmiccai} on the INBreast dataset. The total number of pixels is 92,800. The numbers of pixels classified right and wrong for both models are 76,130 and 8,805, respectively. The number of pixels classified right by only using our model is 4,595. The number of pixels classified right by using model~\cite{dhungel2015deepmiccai} is 3,270. We obtain p-value $< 0.001$, which shows our model is significantly better than
model~\cite{dhungel2015deepmiccai}. %$2\times10^{-50}$

\begin{figure}[t]
	\begin{center}
		\begin{minipage}{0.04\linewidth}
			\centerline{\includegraphics[width=0.8cm]{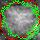}
				\includegraphics[width=0.8cm]{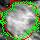}
				\includegraphics[width=0.8cm]{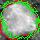}
				\includegraphics[width=0.8cm]{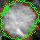}
				\includegraphics[width=0.8cm]{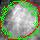}
				\includegraphics[width=0.8cm]{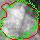}
				\includegraphics[width=0.8cm]{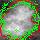}}
		\end{minipage}
		\hspace{6cm}
		\begin{minipage}{0.04\linewidth}
			\centerline{\includegraphics[width=0.8cm]{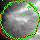}
				\includegraphics[width=0.8cm]{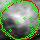}
				\includegraphics[width=0.8cm]{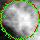}
				\includegraphics[width=0.8cm]{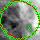}
				\includegraphics[width=0.8cm]{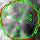}
				\includegraphics[width=0.8cm]{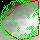}
				\includegraphics[width=0.8cm]{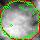}
			}
		\end{minipage}
		\vfill
		\begin{minipage}{0.04\linewidth}
			\centerline{\includegraphics[width=0.8cm]{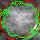}
				\includegraphics[width=0.8cm]{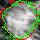}
				\includegraphics[width=0.8cm]{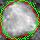}
				\includegraphics[width=0.8cm]{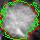}
				\includegraphics[width=0.8cm]{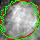}
				\includegraphics[width=0.8cm]{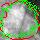}
				\includegraphics[width=0.8cm]{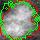}}
		\end{minipage}
		\hspace{6cm}
		\begin{minipage}{0.04\linewidth}
			\centerline{\includegraphics[width=0.8cm]{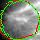}
				\includegraphics[width=0.8cm]{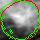}
				\includegraphics[width=0.8cm]{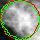}
				\includegraphics[width=0.8cm]{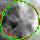}
				\includegraphics[width=0.8cm]{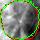}
				\includegraphics[width=0.8cm]{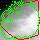}
				\includegraphics[width=0.8cm]{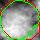}}
		\end{minipage}
		\vfill
		\begin{minipage}{0.04\linewidth}
			\centerline{\includegraphics[width=0.8cm]{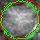}
				\includegraphics[width=0.8cm]{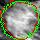}
				\includegraphics[width=0.8cm]{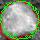}
				\includegraphics[width=0.8cm]{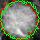}
				\includegraphics[width=0.8cm]{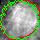}
				\includegraphics[width=0.8cm]{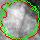}
				\includegraphics[width=0.8cm]{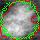}}
		\end{minipage}
		\hspace{6cm}
		\begin{minipage}{0.04\linewidth}
			\centerline{\includegraphics[width=0.8cm]{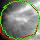}
				\includegraphics[width=0.8cm]{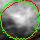}
				\includegraphics[width=0.8cm]{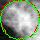}
				\includegraphics[width=0.8cm]{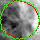}
				\includegraphics[width=0.8cm]{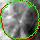}
				\includegraphics[width=0.8cm]{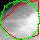}
				\includegraphics[width=0.8cm]{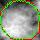}}
		\end{minipage}
		\vfill
		\begin{minipage}{0.04\linewidth}
			\centerline{\includegraphics[width=0.8cm]{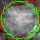}
				\includegraphics[width=0.8cm]{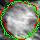}
				\includegraphics[width=0.8cm]{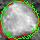}
				\includegraphics[width=0.8cm]{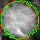}
				\includegraphics[width=0.8cm]{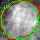}
				\includegraphics[width=0.8cm]{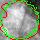}
				\includegraphics[width=0.8cm]{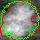}}
		\end{minipage}
				\hspace{6cm}
				\begin{minipage}{0.04\linewidth}
					\centerline{\includegraphics[width=0.8cm]{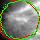}
						\includegraphics[width=0.8cm]{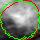}
						\includegraphics[width=0.8cm]{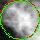}
						\includegraphics[width=0.8cm]{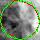}
						\includegraphics[width=0.8cm]{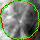}
						\includegraphics[width=0.8cm]{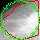}
						\includegraphics[width=0.8cm]{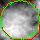}}
				\end{minipage}
		\vfill
		\hspace{0.3cm}
		\begin{minipage}{1\linewidth}
			\center{(a)       \hspace{6cm}         (b)}
		\end{minipage}
		\caption{Visualization of segmentation results using the FCN (the first row), the FCN with adversarial training (the second row), jointly trained FCN-CRF (the third row) and FCN-CRF with adversarial training (the fourth row) on the test sets in the INBreast (a) and DDSM-BCRP (b) datasets. Each column denotes different test samples. Red lines denote the ground truth. The green lines or points denote the segmentation results.}
		\label{fig:all}
	\end{center}
\end{figure}

To further understand the adversarial training, we visualize segmentation results in Fig. \ref{fig:all}. We observe that segmentations in the first row have vague borders and many outliers within the predicted borders. The segmentations in the second row have fewer vague borders and fewer outliers than the predictions in the first row. The results in the last two rows have  sharper and more accurate borders than the first two rows. It demonstrates that the CRF based methods achieves better segmentations on the test sets. The structural learning using CRF eliminates outliers within borders effectively, which makes better segmentation results and more accurately predicted borders. %We will quantize the border prediction in the next paragraph. 

\begin{figure}[t]
	\begin{center}
		%\begin{tabular}{cc}
		\begin{minipage}{0.47\linewidth}
			\centerline{\includegraphics[width=6cm]{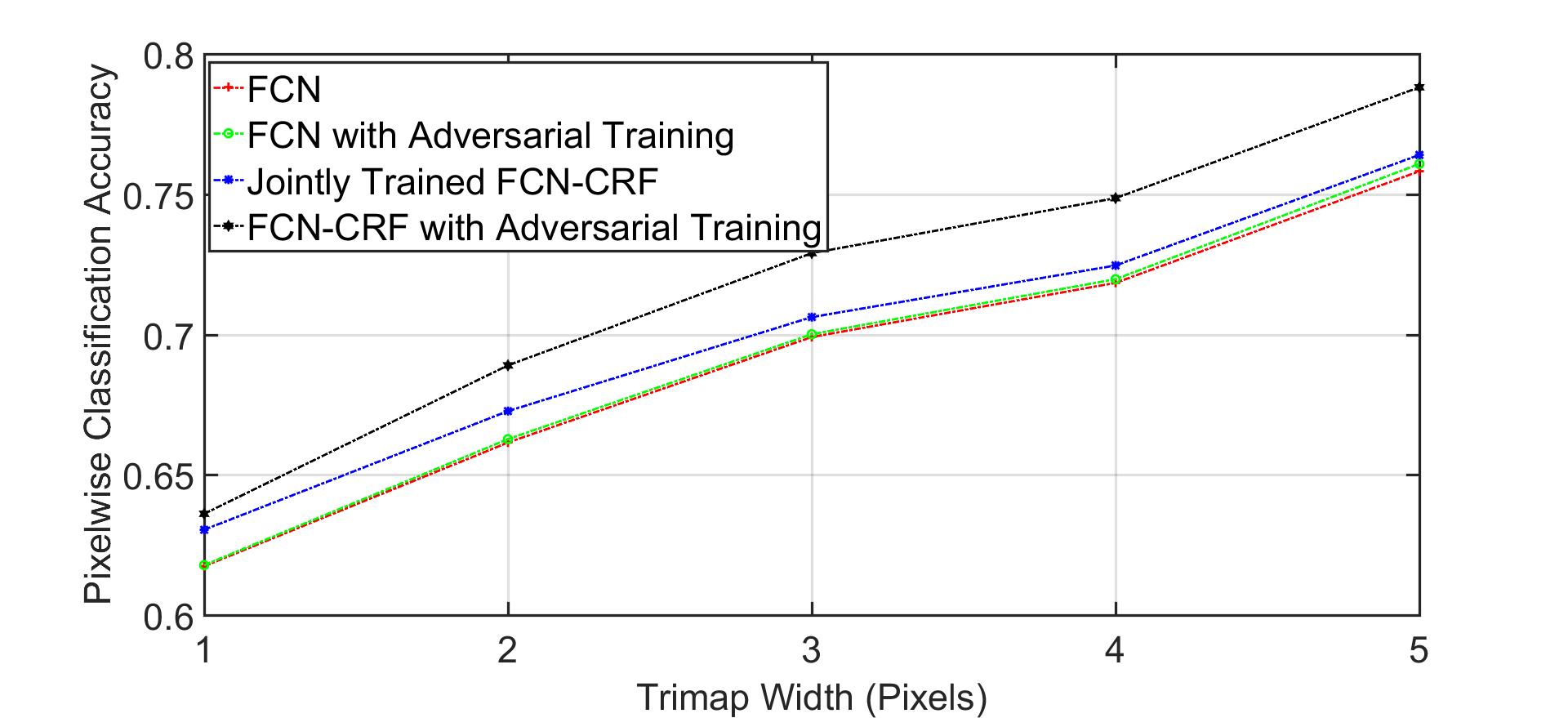}}
			\center{(a)}
		\end{minipage}
		\hspace{0.45cm}
		\begin{minipage}{0.47\linewidth}
			\centerline{\includegraphics[width=6cm]{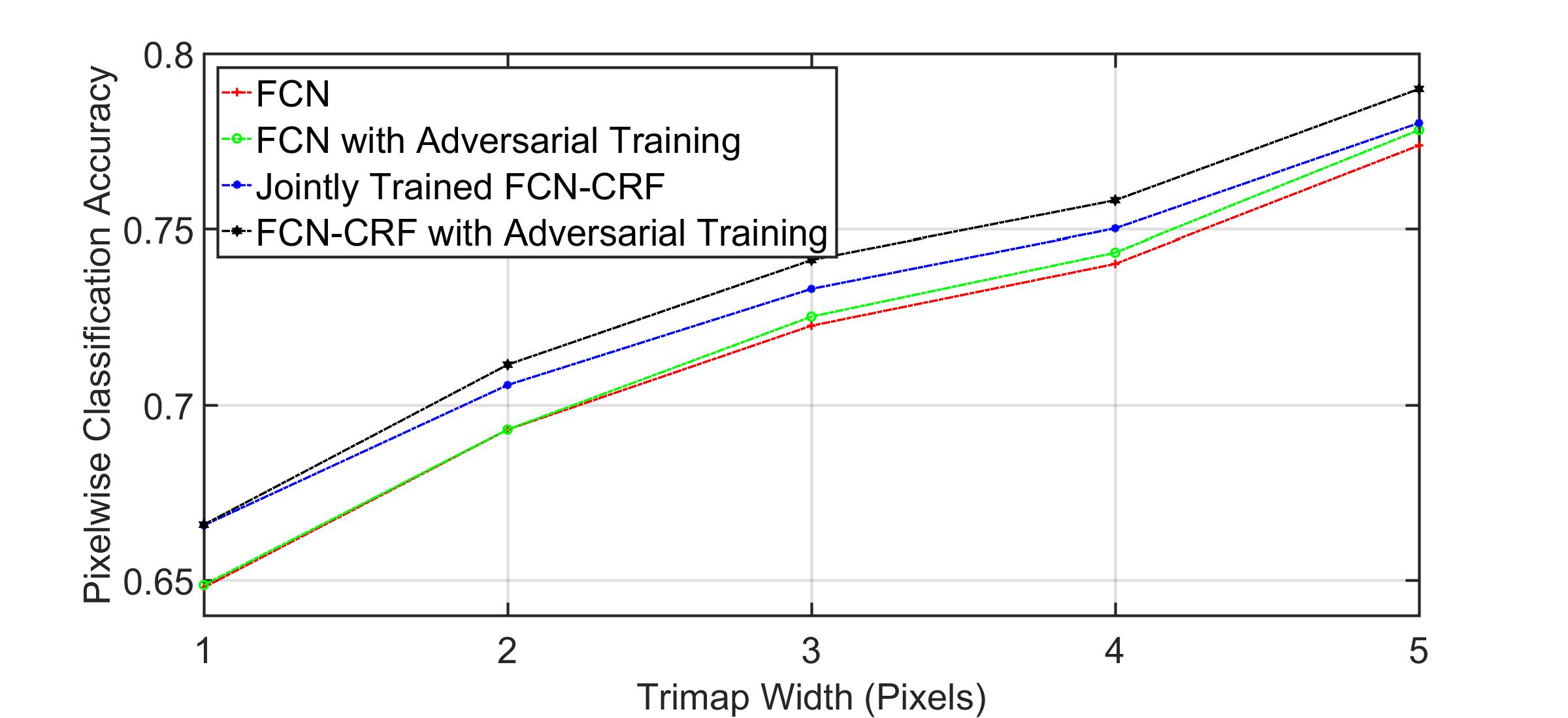}}
			\center{(b)}
		\end{minipage}
		\caption{Accuracy comparisons among FCN, FCN with Adversarial Training, Jointly Trained FCN-CRF and FCN-CRF with Adversarial Training in trimaps with pixel width $1$, $2$, $3$, $4$, $5$ on the Inbreast dataset (a) and the DDSM-BCRP dataset (b). The adversarial training improves the segmentation accuracy on boundaries.}
		\label{fig:trimapinbreast}
	\end{center}
\end{figure}

We further employ the metric based on the trimap to specifically evaluate segmentation accuracy in boundaries \cite{kohli2009robust}. We calculate the accuracies within trimap surrounding the actual mass boundaries (groundtruth) in Fig. \ref{fig:trimapinbreast}. Trimaps on the DDSM-BCRP dataset is visualized in Fig.~\ref{prior}(b). From the figure, accuracies of FCN-CRF with Adversarial Training are 2-3 \% higher than those of FCN-CRF on average and the accuracies of FCN with Adversarial Training are better than those of FCN. The results demonstrate that the adversarial training regularization improves the FCN and FCN-CRF both in the whole image (Dice Index metric) and around the boundaries.
\section{Conclusion}\label{sec:con}
In this paper, we propose an end-to-end trained adversarial FCN-CRF network for mammographic mass segmentation. To integrate the priori distribution of masses and fully explore the power of FCN, a position priori is added to the network. Furthermore, adversarial training is used to handle the small size of training data by reducing over-fitting and increasing robustness. Experimental results demonstrate the state-of-the-art performance of our model on the two most used public mammogram datasets.
{\small
	\bibliographystyle{splncs03}
	\bibliography{aaai}

\begin{thebibliography}{10}
\bibitem{acs}
American cancer society. what are the key statistics about breast cancer?
\end{thebibliography}
}
\end{document}